\documentclass[5p,times]{elsarticle}
\usepackage{lineno}
\modulolinenumbers[5]
\usepackage{amsmath}
\usepackage{amssymb}
\usepackage{bbm} 	
\usepackage{graphicx,caption}
\usepackage{float}
\usepackage{lipsum}
\usepackage{algorithm}
\usepackage{algpseudocode}
\usepackage{natbib}
\usepackage{siunitx}
\usepackage{xcolor}
\usepackage{xparse}
\usepackage{setspace}
\usepackage{soul}
\usepackage{hyperref}
\usepackage{cleveref}

\newcommand{\norm}[1]{\left\lVert#1\right\rVert}
\begin{document}
	\begin{frontmatter}
		\title{ASOCEM: Automatic Segmentation Of Contaminations in cryo-EM}
		\author{Amitay Eldar\corref{cor1}}
		\ead{amitayeldar@tauex.tau.ac.il}
		\author{Ido Amos}
		\ead{idoamos@mail.tau.ac.il}
		\author{Yoel Shkolnisky}
		\ead{yoelsh@tauex.tau.ac.il}
		\cortext[cor1]{Corresponding author}

		\address{Department of Applied Mathematics, School of Mathematical Sciences, Tel-Aviv University, Tel-Aviv ,Israel}

	\begin{abstract}
		 Particle picking is currently a critical step in the cryo-electron microscopy single particle reconstruction pipeline. Contaminations in the acquired micrographs severely degrade the performance of particle pickers, resulting is many ``non-particles'' in the collected stack of particles. In this paper, we present ASOCEM (Automatic Segmentation Of Contaminations in cryo-EM), an automatic method to detect and segment contaminations, which requires as an input only the approximated particle size. In particular, it does not require any parameter tuning nor manual intervention. Our method is based on the observation that the statistical distribution of contaminated regions is different from that of the rest of the micrograph. This nonrestrictive assumption allows to automatically detect various types of contaminations, from the carbon edges of the supporting grid to high contrast blobs of different sizes. We demonstrate the efficiency of our algorithm using various experimental data sets containing various types of contaminations. ASOCEM is integrated as part of the KLT picker \cite{ELDAR2020107473} and is available at  \url{https://github.com/ShkolniskyLab/kltpicker2}.
	\end{abstract}

\end{frontmatter}

\section{Introduction}
Single particle cryo-electron microscopy (cryo-EM) is an established method for high resolution structure determination of macro-molecules. 
A typical cryo-EM data set consists of hundreds of noisy images, called micrographs, with each micrograph containing multiple two-dimensional particle images (that are essentially two-dimensional tomographic projections of the investigated macro-molecule). In order to determine a high resolution three-dimensional  model of the macro-molecule, one needs many thousands of particle images, to overcome their high level of noise. Therefore, one of the first steps towards three-dimensional reconstruction is detecting and segmenting the particles images form the micrographs, a step known as ``particle picking'' \citep{singer2020computational}.

Typical micrographs consist of three types of regions -- regions of particles with added noise, regions of noise only, and regions of contaminations. The latter can cause various problems to existing particle picking algorithms, from high rate of false-positives  (picking ``non-particles'') to completely breaking down the picking process. Over the past few years, two main approaches were developed to address the problem of contaminations in cryo-EM data. The first approach combines contamination detection within the picking process. For pickers based on deep learning, this means including labeled contaminated data as part of the training  process~\citep{PPR:PPR8239,WANG2016325,2016arXiv160505543Z,Tegunov338558}. Other pickers use morphological operators to avoid contaminations~\citep{HEIMOWITZ2018215,LANGLOIS20141}. The second approach separates the task of contamination detection from the picking step. In this approach, one first detects the contaminated regions, and then uses this information (typically in the form of a binary mask) as an input to an existing particle picking algorithm, which has been adapted to skip the contaminated regions. Representatives for this approach include EMHP~\citep{10.1093/bioinformatics/btx500}, which is designed to detect only carbon contaminations through edge detection methods based on Sobel filtering, and MicrographCleaner~\citep{SANCHEZGARCIA2020107498}, which is a deep learning based algorithm that uses a U-net trained on a data set of $539$ manually segmented micrographs.

In this paper, we present ASOCEM: Automatic Segmentation Of  Contaminations in cryo-EM, whose only required parameter is the estimated particle size. Our approach is based on the assumption that the contaminated and uncontaminated regions are Gaussian processes with different and unknown means and covariances. In other words, contaminated and uncontaminated regions have different statistics. This  nonrestrictive assumption allows to detect various types of contaminations, as demonstrated in Section~\ref{sec: experimental results}. We note that our algorithm is also applicable to micrographs with no contaminations, provided the user allows for a low percentage of false positives (that is, uncontaminated regions that are being detected as contaminations). Our algorithm requires a few seconds per micrograph, and returns for each micrograph a binary mask indicating the locations of contaminations that should be excluded by subsequent particle picking. An implementation of our algorithm  has been integrated into the KLT picker~\citep{ELDAR2020107473}, allowing the user to detect contamination (the outputs are binary masks), and also perform particle picking directly from contaminated micrographs. The enhanced KLT picker is available at \url{https://github.com/ShkolniskyLab/kltpicker2}.

\section{Materials and methods}
We model a micrograph as a random discrete function $I: G\rightarrow\mathbb{R}$, where $G\subset[0,1]^2$ is a two-dimensional evenly-spaced  grid. We denote by $\mathcal{G}$ a  partition of the unit square, that is $\mathcal{G}=\left(\mathcal{G}_0,\mathcal{G}_1\right)$ such that 
$\mathcal{G}_0\cap\mathcal{G}_1=\emptyset$ and  $\mathcal{G}_0\cup\mathcal{G}_1=[0,1]^2$. In this notation, $\mathcal{G}_0$ represents the contaminated region and $\mathcal{G}_1$ the uncontaminated region of the micrograph. Note that we use only two types of regions, as a finer partition requires a more complicated model planned as a future research. Denote $G_0 =G\cap \mathcal{G}_{0}$, $G_1=G\cap\mathcal{G}_{1}$, and assume that $I|_{G_0}\sim \mathcal{N}\left(\mu_0,\Sigma_0\right)$, $I|_{G_1}\sim\mathcal{N}\left(\mu_1,\Sigma_1\right)$, where $\mathcal{N}\left(\mu,\Sigma\right)$ is a normal distribution with mean $\mu$ and covariance $\Sigma$. Moreover, assume that $I|_{G_0}$ and $I|_{G_1}$ are independent and  stationary (stationary means that for any two points $x,y\in G_{i}$, $i=1,2$, the covariance $\Sigma_{i}\left(x,y\right)$  depends only on $x-y$).
Given this model, the problem of segmenting a micrograph into ``good'' and ``bad'' regions is stated as follows: estimate $\mathcal{G}$ given a realization of~$I$, where the  parameters $\mu_0$, $\Sigma_0$, $\mu_1$, $\Sigma_1$ are unknown. In practice, a ``realization of~$I$'' is simply the input micrograph. We note that as both regions $ \mathcal{G}_{0},\mathcal{G}_{1} $ are represented by Gaussian processes with unknown parameters, one can't determine, without making  further assumptions, which one represents the contaminated region. To adders this problem, we will assume that the contaminated region has smaller area. In order to estimate $\mathcal{G}$, we maximize the likelihood function $\mathcal{L}\left(\Theta|I\right):= f_I(G;\Theta)$  where $f_{I}$ is the probability density  of~$I$ (to be defined shortly) and $\Theta =\left(\mathcal{G},\mu_0,\Sigma_0,\mu_1,\Sigma_1\right)$. One can show that if the partition $\mathcal{G}$ is not "nice" enough, for example if the boundary of $ \mathcal{G}_0 $ is not smooth or with infinite length, then this maximization problem is ill posed and needs to be regularized~\citep{902291}. We choose a regularization term based on the assumption that the contaminated area and its boundary are not too large,  which yields the maximization problem
\begin{equation}\label{eq:submain_arg_max}
	\hat{\Theta}=\arg\max_\Theta f_I(G;\Theta)\cdot e^{-\left(\alpha\cdot\operatorname{Length}\left(\partial\mathcal{G}_{0}\right)+\beta\cdot\operatorname{Area}\left(\mathcal{G}_{0}\right)\right)},
\end{equation}
where on the right hand side, the left term is the above-mentioned likelihood, the right term is the regularization term with $ \partial\mathcal{G}_{0} $ being the boundary of $ \mathcal{G}_{0} $, and $\alpha,\beta$ are positive parameters. The  mathematical model described in this section (up to equation \eqref{eq:submain_arg_max} ), is a generalization of the well known Chan-Vese model~\citep{902291} to the case where the contaminated and uncontaminated regions are assumed to be Gaussian processes with different and unknown mean and covariance. Contrary to our generalization, the classical Chan-Vese model assumes different means and variances, but without any correlations between different points in space.
A common approach to estimate the solution of~\eqref{eq:submain_arg_max} is by alternating maximization, where in each iteration we estimate the solution with respect to one variable while treating the others as constants. Thus, we first  treat $\mathcal{G} $ as constant and estimate the statistical parameters $\mu_{0},\Sigma_{0},\mu_{1},\Sigma_{1}$. Since we have only one realization of $I$ (our input micrograph), we can't estimate the statistical parameters without making further assumptions.  A common assumption in cryo-EM is that the correlation between pixels decays fast as the distance between them grows. To take advantage of this assumption, we partition $[0,1]^2$ into $N$ non-overlapping squares $\mathcal{B}_i$, and assume for simplicity that all squares have the same area $\mathcal{B}$. Denote $B_i = \mathcal{B}_i \cap G$, and  assume that pixels from different squares are  independent, which implies that for all  $B_i$ and $B_j$,  
\begin{equation}\label{eq:region independence}
	\left(f_{I}\right)|_{B_i\cup B_j}=\left(f_{I}\right)|_{B_i}\cdot\left(f_{I}\right)|_{B_j}.
\end{equation}
Since $I|_{G_0}$ is stationary, for each $B_i,B_j\subset G_0$ we have that $\Sigma_{0}|_{B_i} =\Sigma_{0}|_{B_j}$,  which we denote by $\Sigma^B_0 := \Sigma_{0}|_{B_i} $. The latter implies that the density of pixel intensities of~$I$ on each $B_i\subset G_0$ can be written as  
\begin{equation*}
	\rho^{0}_{B_i} := f_{I} |_{B_i} = \frac{\exp{\left(-\frac{1}{2} \left(x_{B_i}-\mu_0\right)^T\left(\Sigma_{0}^{B}\right)^{-1}\left(x_{B_i}-\mu_0\right)\right)}}{\sqrt{\left(2\pi\right)^n\det{\Sigma^B_0}}},
\end{equation*}
where $x_{B_i}\in \mathbb{R}^n$ is the column stack vector of $B_i$. For $I|_{G_1}$ we define $\rho^1_{B_i}$ in a similar way. Next, denote
\begin{equation}\label{eq:boundery}
	f_{B_{i}}^{0}=
	\left\{
	\begin{array}{cl}
		\rho^0_{B_i}  & B_i\cap G_0\neq\emptyset, \\
		1 & \mbox{otherwise,} 
	\end{array}
	\right.
\end{equation}
then, by ignoring the error induced by boxes $\mathcal{B}_{i}$ on the boundary of  $\mathcal{G}_{0}$ and using~\eqref{eq:region independence}, we have that 
\begin{equation}\label{eq:density product}
	f_{I|_{G_0}}=\prod^{N}_{i=1}f_{B_{i}}^{0}.
\end{equation}
The same derivation for $\mathcal{G}_{1}$  yields $f_{I}|_{G_1}=\prod^{N}_{i=1}f_{B_{i}}^{1}$,	where $f_{B_{i}}^{1}$ is defined similarly to~\eqref{eq:boundery} but on $G_1$. As mentioned above, $I|_{G_{0}}$ and $I|_{G_{1}}$ are independent, and therefore, using~\eqref{eq:density product} we get that $ f_I $ from \eqref{eq:submain_arg_max} is given by
\begin{equation*} 
	f_I = f_I |_{G_{0}}\cdot f_I |_{G_{1}} = \prod^{N}_{i=1}f_{B_{i}}^{0} \cdot \prod^{N}_{j=1}f_{B_{j}}^{1}.
\end{equation*} 
Substituting the latter equation back into~\eqref{eq:submain_arg_max} and taking the logarithm gives 
\begin{equation}\label{main_arg_max}
\begin{split}
	\hat{\Theta}=\arg\max_{\Theta} \sum_{i=1}^{N}&\log{f_{B_{i}}^{0}}+\sum_{j=1}^{N}\log{f_{B_{j}}^{1}}\\ &-\Big(\alpha\cdot\operatorname{Length}\left(\partial\mathcal{G}_{0}\right)+\beta\cdot\operatorname{Area}\left(\mathcal{G}_{0}\right)\Big).
\end{split}
\end{equation} 
By differentiating~\eqref{main_arg_max} with respect to $\mu_0$ and $\Sigma^B_0$ and equating to zero, we get that these parameters can be estimated as the sample mean and sample covariance of $\left\{x_{B_i}\big|B_i\cap G_0\neq\emptyset\right\}$  (see~\citep{ANDERSON1985147} for more details). The parameters $\mu_1,\Sigma^B_1$ are estimated similarly. 

Next, treating $\mu_0$, $\Sigma^B_0$, $\mu_1$, $\Sigma^B_1$ as constants, we turn to estimating $\mathcal{G}$, which is the partition of the unit square to the contaminated and uncontaminated regions. Inspired by the level set method of Chan-Vese~\citep{902291}, instead of estimating $\mathcal{G}$, we estimate a  level set Lipschitz function $\varphi:[0,1]^2\rightarrow\mathbb{R}$ such that $\{\varphi>0\}=\mathcal{G}_0,\;\{\varphi<0\}=\mathcal{G}_1,\;\{\varphi=0\}=\partial\mathcal{G}_0$.  To that end, we rewrite~\eqref{main_arg_max} as an integral maximization problem as follows. Using the Heaviside step function 
\begin{equation}
	H(x)=
	\left\{
	\begin{array}{ll}
		1  & x>0, \\
		0 & \text{otherwise,} 
	\end{array}
	\right.
\end{equation}
and its derivative the Dirac $\delta$ function, it holds that 
\begin{equation}\label{length and area to integral}
	\begin{aligned}
		\operatorname{Area}\left(\mathcal{G}_{0}\right) &= \int_{[0,1]^2}H(\varphi(x))dx,\\
		\operatorname{Length}\left(\partial\mathcal{G}_{0}\right)&=\int_{[0,1]^2}\delta(\varphi(x))\norm{\nabla\varphi(x)}dx.
	\end{aligned}
\end{equation}
In addition, for every $B_i \subset G_{0}$ or $B_i\cap G_{0}=\emptyset$,  it follows from the definition of $f^0_{B_i}$ (see \eqref{eq:boundery}) that
\begin{equation*}
	\log{f^0_{B_i}} = \frac{1}{\mathcal{B}}\int_{[0,1]^2}\log{f^0_{B_i}}\cdot\mathbbm{1}_{\mathcal{B}_i}(x)\cdot H(\varphi(x)) dx,
\end{equation*}
where $ \mathbbm{1}_{\mathcal{B}_i} $ is the indicator function of the set $ \mathcal{B}_i$.
Similarly, for every $B_i \subset G_{1}$ or $B_i\cap G_{1}=\emptyset$
\begin{equation*}
	\log{f^1_{B_i}} = \frac{1}{\mathcal{B}}\int_{[0,1]^2}\log{f^1_{B_i}}\cdot\mathbbm{1}_{\mathcal{B}_i}(x)\cdot \left(1-H(\varphi(x))\right) dx.
\end{equation*}
By ignoring as above the error induced by boxes $\mathcal{B}_{i}$ on the boundary of $\mathcal{G}_{0}$, the latter two equations imply that the sums of the logarithms in \eqref{main_arg_max} can be written as 
\begin{equation}\label{eq: log to integral}
\begin{aligned}
			\sum_{i=1}^{N}\log{f^0_{B_i}}&=\frac{1}{\mathcal{B}}\int_{[0,1]^2}\sum_{i=1}^{N}\log{f^0_{B_i}}\cdot\mathbbm{1}_{\mathcal{B}_i}(x)\cdot H(\varphi(x))dx,\\
			\sum_{i=1}^{N}\log{f^1_{B_i}}&=\frac{1}{\mathcal{B}}\int_{[0,1]^2}\sum_{i=1}^{N}\log{f^1_{B_i}}\cdot\mathbbm{1}_{\mathcal{B}_i}(x)\cdot \left(1-H(\varphi(x))\right) dx.
\end{aligned}
\end{equation}
Using~\eqref{length and area to integral},~\eqref{eq: log to integral} and the notation
\begin{equation*}
	g_0(x)=\sum_{i=1}^{N}\log{f^0_{B_i}}\cdot\mathbbm{1}_{\mathcal{B}_i}(x),\quad g_1(x)=\sum_{i=1}^{N}\log{f^1_{B_i}}\cdot\mathbbm{1}_{\mathcal{B}_i}(x),
\end{equation*}
and recalling that $\mu_{0},\Sigma_{0}^{B},\mu_{1},\Sigma_{1}^{B}$ are fixed, the optimization problem~\eqref{main_arg_max} becomes
\begin{equation}\label{eq: maximization problem for phi}
	\begin{split}
		\hat{\varphi}=\arg\max_{\Theta} \frac{1}{\mathcal{B}}\int_{[0,1]^2} g_{0}(x)H(\varphi(x)) + g_{1}(x)\left(1-H(\varphi(x))\right)\\
		-\beta \mathcal{B}\cdot H(\varphi(x))-\alpha \mathcal{B}\cdot\delta(\varphi(x))\norm{\nabla\varphi(x)}dx. 
	\end{split}
\end{equation}
Using calculus of variations, we show in~\ref{appendix: argmax derivation} that the  solution $\hat{\varphi}$ of the latter optimization problem is given as the limit $\hat{\varphi}(x)=\lim_{t\rightarrow\infty}\bar{\varphi}(x,t)$, where $\bar{\varphi}(x,t)$ is the solution of the differential equation
\begin{equation}\label{eq:differential_eq}
	\bar{\varphi}_t = \delta_a(\bar{\varphi})\cdot\Bigg[-\alpha\cdot\text{div}\left(\frac{\nabla\bar{\varphi}}{\norm{\nabla\bar{\varphi}}}\right)-\beta+\frac{1}{\mathcal{B}}\sum_{i=1}^N\mathbbm{1}_{\mathcal{B}_{i}}\log{\left(\frac{f^0_{B_i}}{f^1_{B_i}}\right)}\Bigg],
\end{equation}
with $\delta_a$ being a smooth function which converges to the Dirac~$\delta$ function when $ a\rightarrow0$ (see Appendix \eqref{eq:regularization for step and delta} for details).
We estimate the solution of~\eqref{eq:differential_eq} using a finite difference scheme as proposed in~\citep{902291}.

\section{Experimental  results}\label{sec: experimental results}
	In this section, we demonstrate the performance of our ASOCEM algorithm on various  data sets. We also demonstrate that integrating the algorithm with the KLT picker~\citep{ELDAR2020107473} improves particle picking results. We compare the performance of our algorithm to that of MicrographCleaner~\citep{SANCHEZGARCIA2020107498} as both algorithms are fully automatic and only require as a parameter an estimate of the particle size (with MicrographCleaner we used the default detection threshold of $0.2 $). The performance measures we report for the algorithms are sensitivity and specificity~\citep{GADDIS1990591}. In our settings, sensitivity is the ratio between the number of pixels an algorithm detects as contamination and the total number of pixels labeled as contamination. Specificity is the ratio between the number of pixels detected by the algorithm as uncontaminated and the total number of uncontaminated pixels in the micrograph. We would like both sensitivity and specificity to be as close to one as possible. We tested our algorithm using three data sets, described in detail below. The first data set was provided by the MicrographCleaner team, and was used as part of the MicrographCleaner training data. The two other data sets were annotated by our team. The main difference between the first data set and the other two is that the latter have not been used in the training process of MicrographCleaner, and therefore, provide more objective performance measures.
	
	Finally we demonstrate how ASOCEM may improve the picking results of the KLT picker~\citep{ELDAR2020107473}. It is worth noting that the popular RELION \citep{scheres2015semi} and EMAN \citep{Ludtke} particle pickers  do not support the use of contamination masks, so we cannot use them to demonstrate the benefits of ASOCEM. 
	\subsection{MicrographCleaner  data}
\begin{figure}
	\includegraphics[width=1\linewidth]{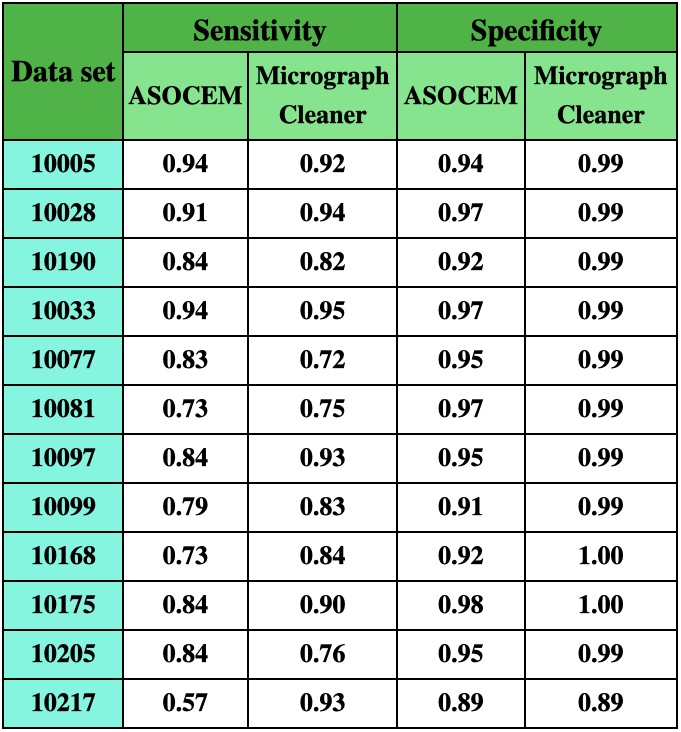}
	\centering
	\caption{Average performance using the MicrographCleaner training data set. The data set consists of $ 111 $ micrographs of different sizes and with various contaminations, extracted from $12$ cryo-EM data sets from the EMPIAR repository~\citep{empiar}. The names of the data sets are given in the left column.}
	\label{fig:microTable}
\end{figure}

\begin{figure}
	\includegraphics[width=1\linewidth]{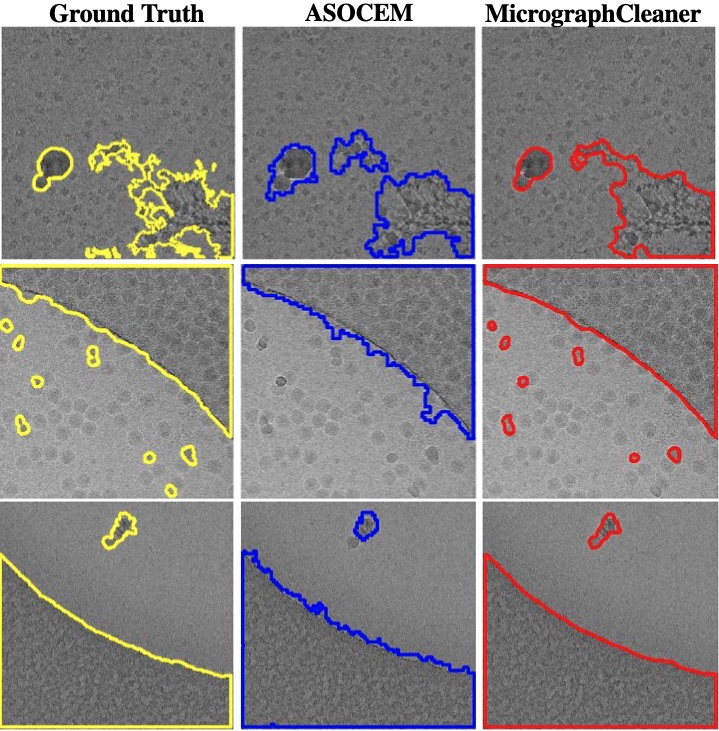}
	\centering
	\caption{Examples of detection results for $3$ micrographs from the MicrographCleaner data set for both ASOCEM (middle column) and MicrographCleaner (right column). Manually annotated ground-truth is shown on the left column.}
	\label{fig:microMic}
\end{figure}

	The MicrographCleaner data set consists of $111$ micrographs, of different sizes and with various contaminations, extracted from 12 different cryo-EM data sets. These micrographs were used in the training process of the MicrographCleaner algorithm. The ground-truth for this data set was generated by manual annotation by the MicrographCleaner team. 
	Table~\ref{fig:microTable} presents the average performance measures for each of the data sets. Figure~\ref{fig:microMic} shows $3$ micrographs and the detection results of ASOCEM and MicrographCleaner.
	
	\subsection{Untrained data}
\begin{figure}
	\includegraphics[width=1\linewidth]{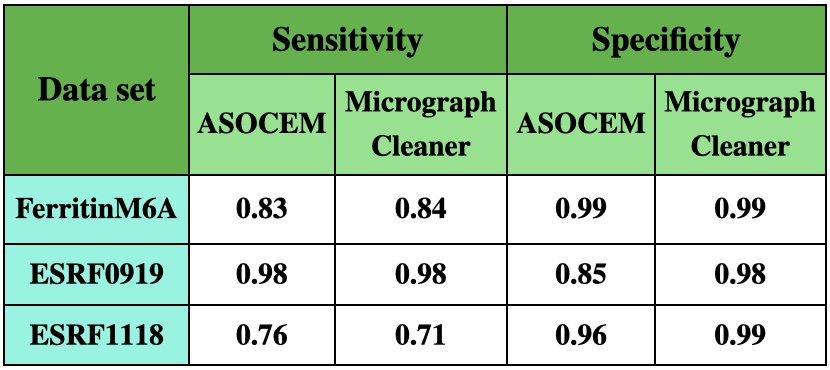}
	\centering
	\caption{Average performance for the untrained data set. This data set consists of $13$ micrographs, which were extracted from $3$ different cryo-EM data sets \citep{GabiFrank,10.1093/pcp/pcab046,CASPY2020148253}.}
	\label{fig:idoCaspiTable}
\end{figure}

\begin{figure}
	\includegraphics[width=1\linewidth]{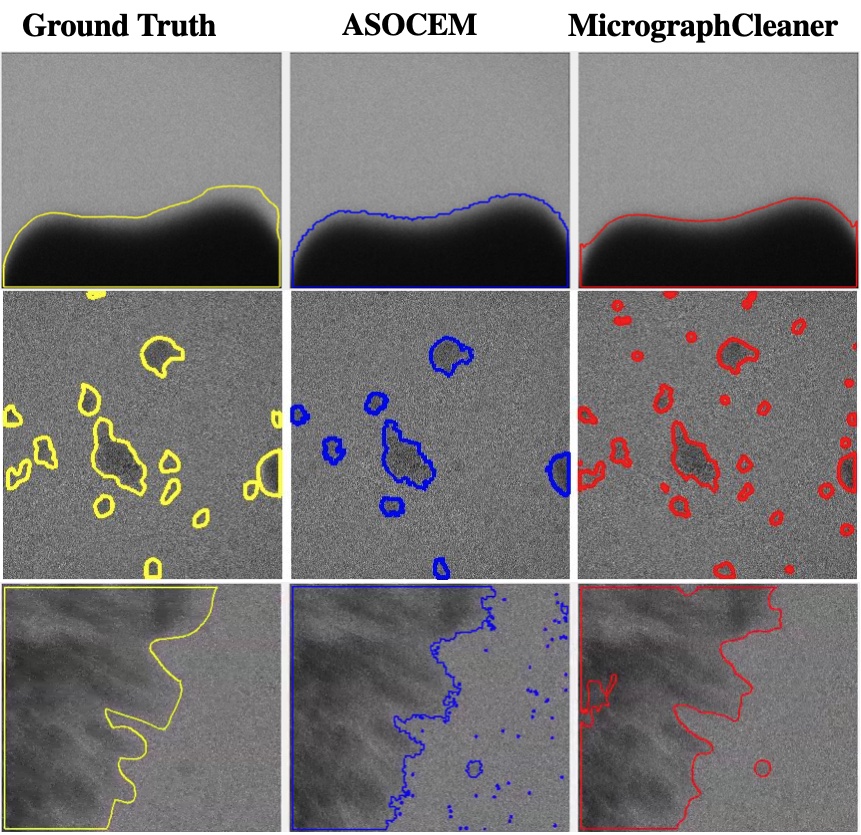}
	\centering
	\caption{Examples of detection results for $3$ micrographs from the  untrained data set for both ASOCEM (middle column) and MicrographCleaner (right column). Manually annotated ground-truth is shown on the left column. The top-to-bottom micrographs were taken, by the same order, from the following ~cryo-EM data sets  \citep{GabiFrank,10.1093/pcp/pcab046,CASPY2020148253}.}
	\label{fig:idoCaspiMic}
\end{figure}

	This test data consists of $13$ micrographs with various contaminations, extracted from~$3$ different cryo-EM data sets \citep{GabiFrank,10.1093/pcp/pcab046,CASPY2020148253}, $ 8 $ micrograph from the first data set, $ 3 $ from the second and~$ 2 $ from the third. The ground truth for this test data was generated by manual annotation. All micrographs were downsampled to size $ 800\times800 $ pixels prior tp contamination detection.  Table~\ref{fig:idoCaspiTable} shows the average performance measures for these test data. Figure~\ref{fig:idoCaspiMic} shows $3$ micrographs (one from each data set) and the detection results of ASOCEM and MicrographCleaner.
	
	\subsection{Integrating ASOCEM with the KLT picker}
\begin{figure*}
	\includegraphics[width=1\linewidth]{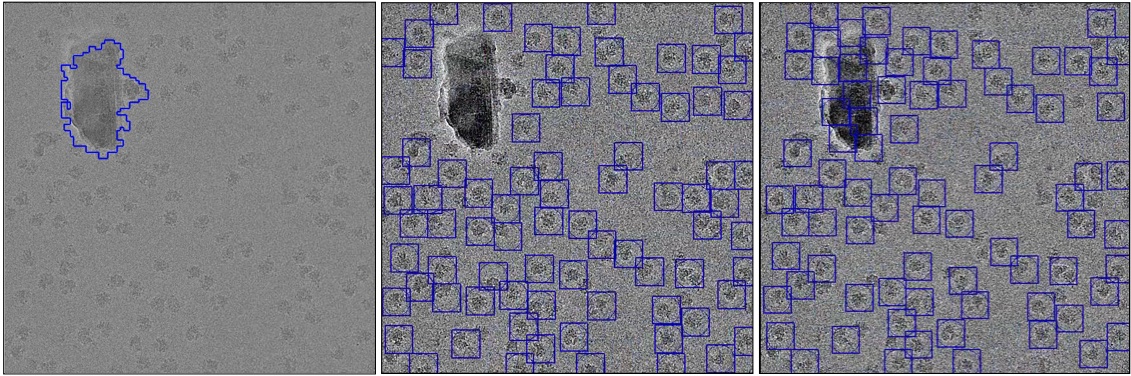}
	\centering
	\caption{Integrating contamination detection with particle picking. Left column shows the contamination detected by the ASOCEM algorithm. Middle column shows the picking results when combining the ASOCEM algorithm with the KLT picker~\citep{ELDAR2020107473}. Right column shows the picking results of the KLT picker when ignoring the presence of contaminations. This example demonstrates the case where the KLT picker is picking contaminated data as particles, but still manages to pick "good" particles as well. The micrograph in this example is part of the Plasmodium falciparum 80S ribosome data set~\citep{PMID:24913268}.}
	\label{fig:kltPickerMic50}
\end{figure*}

\begin{figure*}
	\includegraphics[width=1\linewidth]{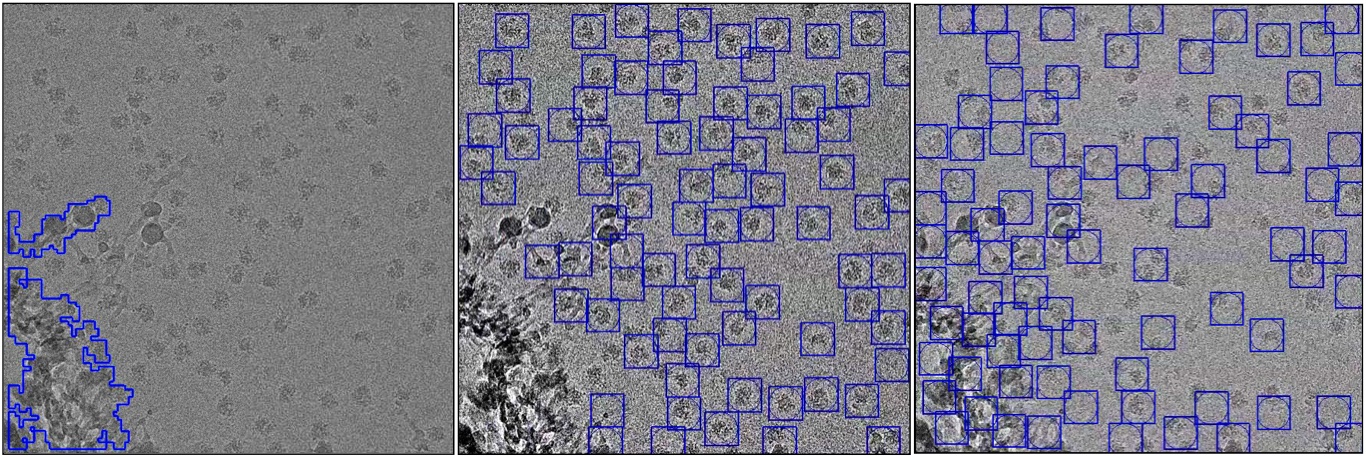}
	\centering
	\caption{Same experiment as in Figure~\ref{fig:kltPickerMic50} but for a different micrograph. For this micrograph, ignoring the presence of contaminations during particle picking results in failure of the particle picking algorithm. The micrograph in this example is part of the Plasmodium falciparum 80S ribosome data set~\citep{PMID:24913268}.}
	\label{fig:kltPickerMic6}
\end{figure*}

		In this section, we demonstrate that integrating contamination detection with particle picking may improve the results of the picking. To demonstrate this point, we compare between the performance of the KLT picker~\citep{ELDAR2020107473} and an enhanced version of the KLT picker into which the ASOCEM algorithm has been integrated. In the enhanced KLT picker, the ASOCEM algorithm first detects all contaminated pixels, and then the KLT picker ignores these pixels during picking. The KLT picker with integrated ASOCEM algorithm is available at \url{https://github.com/ShkolniskyLab/kltpicker2}. To compare both versions of the KLT picker we use micrographs from the the Plasmodium falciparum 80S ribosome data set~\citep{PMID:24913268}. Figure~\ref{fig:kltPickerMic50} shows a micrograph for which the KLT picker is picking contaminated pixels as particles, but still manages to pick ``good'' particles as well. The micrograph in Figure~\ref{fig:kltPickerMic6} shows a worse and more common case where contamination results in failure of the picking process. In both cases, the picking results of the enhanced algorithm are much better.
	  
\section{Discussion}
		Before comparing the detection results of ASOCEM and MicrographCleaner, 	
		it is important to note that the latter allows the user to tune the detection threshold of the algorithm. Nevertheless, we used the default threshold of~$0.2$ in all experiments. In some data sets, changing the threshold  may improve the results, however, as noted in~\citep{SANCHEZGARCIA2020107498}, the cryo-EM field is moving towards streaming and automatic processing, and thus, default parameters should perform decently in most cases. As of running times, both algorithms have comparable running times, of about a few seconds per micrograph.
\subsection{MicrographCleaner data}
It is not surprising that for this data MicrographCleaner often slightly outperforms ASOCEM, as this data was used to train the MicrographCleaner algorithm. The sensitivity results of both algorithms are very similar, but MicrographCleaner has  slightly better specificity scores (though both algorithms yield very high scores). From the user's perspective, this means that both algorithms will detect the same high percentage of contaminations, while ASOCEM leaves slightly less data for subsequent particle picking. It is important to note that all micrographs in this data set are only available after being downsampled by the MicrographCleaner team to size of about $ 400\times400 $ pixels. Such sizes are sub-optimal for the ASOCEM algorithm that performs best when the micrographs are at least of size $ 600\times600 $ pixels. The reason ASOCEM benefits from larger micrographs is due to the covariances estimation step (see  the paragraph below equation~ \eqref{eq:submain_arg_max}). In this step, the contaminated and uncontaminated regions are being divided into non-overlapping boxes of the same size, which are being used to approximate the covariance of each region.
In practice, each box is of size $ 25\times 25 $ pixels and one needs enough of them in order to get a good approximation. As the number of boxes depends on the area of the contamination, it seems that downsampling the micrograph to less then $ 600\times 600 $ pixels is sub-optimal in most cases.  

\subsection{Untrained data}
This data best represents a fair comparison between ASOCEM and MicrographCleaner, as it contains micrographs which have not been used to train the MicrographCleaner algorithm. The results of both algorithms are comparable, with a slight advantage to ASOCEM on the sensitivity measure, and a slight advantage to MicrographCleaner on the specificity measure.

\section{Conclusion}
	In this paper, we presented ASOCEM, an automatic algorithm for detecting contaminations in cryo-EM micrographs. The algorithm is inspired by the well known Chan-Vese algorithm and is based on a general statistical model that allows it to detect various types of contaminations. To demonstrate the performance of our algorithm, we tested ASOCEM on various data sets, showing performance that is comparable to that of MicrographCleaner, but without requiring any training,  which is labor-intensive as it requires to manually annotate contaminations in a large set of micrographs. 
	
	The ASOCEM algorithm partitions the micrographs into two regions: contaminated and uncontaminated. This partition is necessarily sub-optimal, as it may happen that the statistical properties of particle regions resemble more to those of contamination regions than to those of noise-only regions. In such cases, particles would be classified as contaminations. ASOCEM alleviates this problem by computing the area of  each region marked as contamination and keeping only those regions whose areas are significantly larger than the particle size. In fact, this is the only place where the particle size is used by the algorithm. A better approach would be to partition the micrographs into three different regions: contaminations, particle plus noise, and noise only. This approach will eliminate the need to provide the algorithm with the approximate particle size, turning the algorithm into completely parameter free. 

\section*{Acknowledgments}
This research was supported by the European Research Council (ERC) under the European Union's Horizon 2020 research and innovation programme (grant
agreement 723991 - CRYOMATH).

	\appendix
	\section{Deriving the solution of \eqref{eq: maximization problem for phi} }\label{appendix: argmax derivation}
	Denoting
	\begin{align}
			F(\varphi)=& \frac{1}{\mathcal{B}}\int_{[0,1]^2} g_{0}(x)H(\varphi(x)) + g_{1}(x)\left(1-H(\varphi(x))\right)\\\nonumber
			&-\beta \mathcal{B}\cdot H(\varphi(x))-\alpha \mathcal{B}\cdot\delta(\varphi(x))\norm{\nabla\varphi(x)}dx,
	\end{align}
	equation \eqref{eq: maximization problem for phi} becomes
	\begin{equation}\label{eq: functional maximization}
		\hat{\varphi}=\arg\max_{\varphi}F(\varphi),
	\end{equation}
	where $ \varphi $ is a Lipschitz function. In\citep[p.45]{vese2015variational},  it is shown that the optimum $ \hat{\varphi} $ of~\eqref{eq: functional maximization} is given as the limit $\hat{\varphi}(x)=\lim_{t\rightarrow\infty}\bar{\varphi}(x,t)$, where $ \bar{\varphi}(x,t) $ is the solution of the gradient descent equations
	\begin{equation}\label{eq:differential eq not regulerizied}
	\begin{aligned}
		\frac{d\bar{\varphi}}{dt} &= \frac{dF}{d\bar{\varphi}}\\
		\bar{\varphi}(0,x)&=\bar{\varphi}_{0}(x)\\
		\frac{\delta\left(	\bar{\varphi}\right)}{\norm{\nabla\bar{\varphi}}}\cdot\frac{\partial\bar{\varphi}}{\partial\vec{n}}&=0 \text{ on } \partial\left([0,1]^2\right),
	\end{aligned}
 	\end{equation}
	where $ \frac{dF}{d\bar{\varphi}} $ is the 
	 functional derivative~\citep[p.27]{CalculusofVariations},  $ \bar{\varphi}_{0}(x)$ is an initial condition \big(in our case we used a spherical cap $ \bar{\varphi}_0(x,y) = 0.25- \big((x-0.5)^2+(y-0.5)^2\big) $\big), and  $ \frac{\partial}{\partial\vec{n}} $ is the normal derivative with respect to the boundary of $ [0,1]^2 $. It is well known that $ \frac{dF}{d\bar{\varphi}}$ can be deduced from the total derivative $\frac{d}{d\epsilon} \big|_{\epsilon=0}F(\bar{\varphi}+\epsilon\cdot u) $ ~\citep[p.14]{CalculusofVariations},
	where $\epsilon $ is a real number and $ u:[0,1]^2 \rightarrow\mathbb{R}$ is an arbitrary Lipschitz function with the same boundary condition as $ \bar{\varphi} $, namely $\frac{\delta\left(	u\right)}{\norm{\nabla u}}\cdot\frac{\partial u}{\partial\vec{n}} = 0 $ . Therefore,  to compute $ \frac{dF}{d\bar{\varphi}} $  we turn to compute $ \frac{d}{d\epsilon} \big|_{\epsilon=0}F(\bar{\varphi}+\epsilon\cdot u) $.  As $ \frac{d}{d\epsilon}\big|_{\epsilon=0} $ is a linear operator, we can differentiate each of the terms in $ F(\bar{\varphi}) $ separately. The first term is 
	\begin{align}\label{eq:1 term}
		\nonumber&\frac{d}{d\epsilon}\bigg|_{\epsilon=0}\;\frac{1}{\mathcal{B}}\int_{[0,1]^2} g_{0}(x)H\big(\bar{\varphi}(x)+\epsilon\cdot u(x)\big)dx\\\nonumber
		&=\frac{1}{\mathcal{B}}\int_{[0,1]^2} g_{0}(x)\frac{d}{d\epsilon}\bigg|_{\epsilon=0}H\big(\bar{\varphi}(x)+\epsilon\cdot u(x)\big)dx\\
		&=\frac{1}{\mathcal{B}}\int_{[0,1]^2}g_{0}(x)\delta(\bar{\varphi}(x))u(x)dx.
	\end{align}
	For simplicity, in subsequent derivations, we drop the dependence on $ x $.
	The second term is
	\begin{align}\label{eq:2 term}
		\nonumber&\frac{d}{d\epsilon}\bigg|_{\epsilon=0}\;\frac{1}{\mathcal{B}}\int_{[0,1]^2} g_{1}\bigg(1-H\big(\bar{\varphi}+\epsilon\cdot u\big)\bigg)dx\\\nonumber
		&=\frac{1}{\mathcal{B}}\int_{[0,1]^2} g_{1}\frac{d}{d\epsilon}\bigg|_{\epsilon=0}\bigg(1-H\big(\bar{\varphi}+\epsilon\cdot u\big)\bigg)dx\\
		&=-\frac{1}{\mathcal{B}}\int_{[0,1]^2}g_{1}\delta(\bar{\varphi})udx.
	\end{align}
	The third term is
	\begin{align}\label{eq:3 term}
	\nonumber	&\frac{d}{d\epsilon}\bigg|_{\epsilon=0}\;-\int_{[0,1]^2} \beta\cdot H\big(\bar{\varphi}+\epsilon\cdot u\big)dx\\\nonumber
		&=-\int_{[0,1]^2} \beta\cdot\frac{d}{d\epsilon}\bigg|_{\epsilon=0} H\big(\bar{\varphi}+\epsilon\cdot u\big)dx\\
		&=-\int_{[0,1]^2}\beta\cdot\delta(\bar{\varphi})udx.
	\end{align}
	Finally, the forth term is
	\begin{align}
		\nonumber&\frac{d}{d\epsilon}\bigg|_{\epsilon=0}-\int_{[0,1]^2} \alpha\cdot\delta(\bar{\varphi}+\epsilon\cdot u)
		\norm{\nabla\bar{\varphi}+\epsilon\cdot u} dx \\\nonumber
		&=-\int_{[0,1]^2} \alpha\cdot\frac{d}{d\epsilon}\bigg|_{\epsilon=0}\delta(\bar{\varphi}+\epsilon\cdot u)\cdot\norm{\nabla\bar{\varphi}+\epsilon\cdot u} dx  \\
		&=-\int_{[0,1]^2}\alpha\delta(\bar{\varphi})\cdot\text{div}\left(\frac{\nabla\bar{\varphi}}{\norm{\nabla\bar{\varphi}}}\right)udx, \label{eq:4 term}
	\end{align}
	where the last step \eqref{eq:4 term} involves basic differential calculus,  similar to the one in \citep{902291} (we omit the calculations for the sake of brevity). 
	The desired functional derivative is equal to the integrand of the sum of \eqref{eq:1 term}--\eqref{eq:4 term} after dropping the multiplication by $ u(x) $, that is
	\begin{equation}\label{eq: functional derivative}
		\frac{dF}{d\bar{\varphi}} =\delta(\bar{\varphi})\cdot\Bigg[-\alpha\cdot\text{div}\left(\frac{\nabla\bar{\varphi}}{\norm{\nabla\bar{\varphi}}}\right)-\beta+\frac{1}{\mathcal{B}}\sum_{i=1}^N\mathbbm{1}_{\mathcal{B}_{i}}\log{\left(\frac{f^0_{B_i}}{f^1_{B_i}}\right)}\Bigg].
	\end{equation}
  In order to estimate the solution of  \eqref{eq:differential eq not regulerizied}, we follow Chan-Vese method \citep{902291}, and  replace $ H$ and $\delta $ by their regularized versions
	\begin{align}\label{eq:regularization for step and delta}
		H_a(z)&=\frac{1}{2}\left(1+\frac{2}{\pi}\arctan(za^{-1})\right),\\\nonumber
		\delta_a(z)&=\frac{d}{dz}H_a(z)= \frac{a}{\pi\left(a^2+x^2\right)},
	\end{align} 
	which means practically replacing $ \delta $ with $ \delta_a $ in \eqref{eq: functional derivative}. Then, we use the same finite differences scheme and the same parameters $ a=dt=1 $ as in \citep{902291} to estimate the solution of the differential equation \eqref{eq:differential eq not regulerizied}.   
\bibliography{asocemBib}
\end{document}